 \documentclass[letterpaper, 10 pt, conference]{ieeeconf}  

\usepackage{graphics} 
\usepackage{epsfig} 
\usepackage{amsmath} 
\usepackage{amssymb}  
\usepackage{algorithm}
\usepackage{mathrsfs}
\usepackage{mathtools}
\usepackage{multirow}
\usepackage{comment}
\usepackage{color}
\usepackage{cite}
\usepackage{url}
\usepackage{soul}
\usepackage{listings}
\usepackage[utf8]{inputenc}
\usepackage[T1]{fontenc}
\usepackage[english]{babel}
\usepackage{xcolor}
\usepackage[normalem]{ulem}
\usepackage{xpatch} 

\newtheorem{theorem}{Theorem}

\usepackage{graphicx}
\usepackage{graphics}
\usepackage{amssymb}
\usepackage{amsmath}
\usepackage{color}
\usepackage{xspace}
\usepackage{algpseudocode}
\usepackage{bbm}
\usepackage{comment}
\usepackage{subfig}
\usepackage{soul}
\usepackage{array}
\usepackage[bottom]{footmisc}
\usepackage{tikz}
\usetikzlibrary{shapes,arrows}
\usetikzlibrary{decorations.pathreplacing,decorations.markings,decorations.pathmorphing,decorations.shapes}
\usetikzlibrary{positioning,fit}
\usetikzlibrary{calc}
\usepackage{multirow}
\usepackage{cancel}
\usepackage{hyperref} 
\usepackage{xr}

\tikzstyle{block}=[draw opacity=0.7,line width=1.4cm]

\definecolor{CranJ}{cmyk}{0,0.69,0.54,0.04} 
\definecolor{PinkJ}{cmyk}{0,0.71,0.43,0.12} 
\definecolor{Cran}{cmyk}{0,0.73,0.41,0.29} 
\definecolor{VRed}{cmyk}{0,0.75,0.25,0.2} 
\definecolor{ORed}{cmyk}{0,0.75,0.75,0} 
\definecolor{CBlue}{cmyk}{1,0.25,0,0} 
\setstcolor{VRed}
\usepackage{bm}



\tikzset{cloud/.pic={
\node[cloud, cloud puffs=10.8,cloud puff arc=110, aspect=2, draw, text width=3cm
    ] () at (0,0) {\tikzpictext};
}}

\author{Jianan Zhu \quad Solmaz S. Kia\\
\normalsize{\emph{University of California Irvine}}
  \thanks{The authors
    are with the Mechanical and Aerospace Eng. Dept. of 
    Univ. of California Irvine, CA 92697,~USA, {\tt\small jiananz1,solmaz@uci.edu}. This work is supported by the U.S. Dept. of Commerce, National Institute of Standards and Technology award 70NANB17H192.
    }%
}

\newcommand{\VV}{\mathcal{V}}

\newcommand{\real}{{\mathbb{R}}}

\newcommand{\argmax}{\operatorname{argmax}}

\newcommand{\prpg}{\mbox{{\footnotesize\textbf{--}}}}
\newcommand{\updt}{\mbox{\textbf{+}}}

\newcommand{\vect}[1]{\boldsymbol{\mathbf{#1}}}
\newcommand{\Bvect}[1]{\bm\bar{\boldsymbol{\mathbf{#1}}}}

\newcommand{\Hvect}[1]{\bm\hat{\boldsymbol{\mathbf{#1}}}}

\newcommand{\oprocendsymbol}{\hbox{$\bullet$}}
\newcommand{\oprocend}{\relax\ifmmode\else\unskip\hfill\fi\oprocendsymbol}

\allowdisplaybreaks

\makeatletter
\renewcommand*{\@opargbegintheorem}[3]{\trivlist
      \item[\hskip \labelsep{\emph{ #1\ #2}}] \emph{(#3):}\ \itshape}
\makeatother

\usepackage{tcolorbox}
\definecolor{mycolor}{rgb}{0.122, 0.435, 0.698}
\makeatletter
\newcommand{\mybox}[1]{%
  \setbox0=\hbox{#1}%
  \setlength{\@tempdima}{\dimexpr\wd0+13pt}%
  \begin{tcolorbox}[colframe=mycolor,boxrule=0.5pt,arc=4pt,
      left=6pt,right=6pt,top=6pt,bottom=6pt,boxsep=0pt,width=\@tempdima]
    #1
  \end{tcolorbox}
}
\makeatother

\usepackage[labelsep=endash]{caption}

\usepackage{cite}

\makeatletter
\def\hlinewd#1{%
\noalign{\ifnum0=`}\fi\hrule \@height #1 \futurelet
\reserved@a\@xhline}
\makeatother

\begin{document}
\makeatletter
\renewcommand*{\@opargbegintheorem}[3]{\trivlist
      \item[\hskip \labelsep{ #1\ #2}] (#3):\ \itshape}
\makeatother
\title{Learning-based Measurement Scheduling for\\ Loosely-Coupled Cooperative Localization}

\author{Jianan Zhu ~~and ~~ Solmaz S. Kia, \emph{Senior Member, IEEE}\\
\normalsize{\emph{University of California Irvine}}
  \thanks{The authors
    are with the Department of Mechanical and Aerospace Engineering,
    University of California Irvine, Irvine, CA 92697, USA, {\tt\small jiananz1@uci.edu,solmaz@uci.edu}. This work was supported by NIST award 70NANB17H192.}%
}

\maketitle

\begin{abstract} 
In cooperative localization, communicating mobile agents use inter-agent relative measurements to improve their dead-reckoning-based global localization. Measurement scheduling enables an agent to decide which subset of available inter-agent relative measurements it should process when its computational resources are limited. Optimal measurement scheduling is an NP-hard combinatorial optimization problem. The so-called sequential greedy (SG) algorithm is a popular suboptimal polynomial-time solution for this problem. However, the merit function evaluation for the SG algorithms requires access to the state estimate vector and error covariance matrix of all the landmark agents (teammates that an agent can take measurements from). This paper proposes a measurement scheduling for CL that follows the SG approach but reduces the communication and computation cost by using a neural network-based surrogate model as a proxy for the SG algorithm's merit function. The significance of this model is that it is driven by local information and only a \emph{scalar} metadata from the landmark agents. This solution addresses the time and memory complexity issues of running the SG algorithm in three ways: (a) reducing the inter-agent communication message size, (b) decreasing
the complexity of function evaluations by using a simpler
surrogate (proxy) function, (c) reducing the required memory size.
Simulations demonstrate our results.
\end{abstract}
\begin{keywords}
\textbf{Keywords}: Localization, multi-agent systems, measurement scheduling, deep neural network (DNN)
\end{keywords}
\section{Introduction}
\label{sec::intro}
\vspace{-0.05in}
In cooperative localization, communicating mobile agents use inter-agent relative measurements to improve their dead-reckoning-based global localization. CL provides an infrastructure-free aiding and creates the opportunity to extend the localization improvement gained by sporadic access to external aidings by a member to the rest of the group. CL is often proposed for applications where access to external landmarks and external aiding signals such as global positioning system (GPS) are challenging, e.g., in underwater operations~\cite{SEW-JMW-LLW-RME:13,bahr2009consistent} or indoor localization for firefighters~\cite{JN-JR-PH-IS-MO-KVSH:14,zhu2018loosely}. However, the implementation of CL is not trivial. By processing inter-agent measurements, strong inter-agent coupling terms are created. Maintaining these inter-agent couplings requires all-to-all communications~\cite{SSK-SF-SM:16}. In the past two decades, to relax this stringent connectivity requirement,  a wide range of decentralized CL algorithms are developed~\cite{roumeliotis2002distributed, SSK-SF-SM:16, carrillo-arce2013decentralized, kia2018serverassisted, POA-CR-RKM:01,HL-FN:13,DM-NO-VC:13,JZ-SSK-TRO:19,LCC-EDN-JLG-SIR:13}. The most relaxed connectivity condition is achieved via loosely coupled CL, in which the correlations are not maintained but accounted for in an implicit manner~\cite{POA-CR-RKM:01,HL-FN:13,DM-NO-VC:13,JZ-SSK-TRO:19,LCC-EDN-JLG-SIR:13,JSR-MY-BDOA-HH-PS:19}. Loosely-coupled CL algorithms require only the two agents involved in a relative measurement to exchange information to process that relative measurement. However, the communication cost is still high if agents take relative measurements from every other agent within their sensing range. The computational cost and the algorithm run-time are also a matter to consider for loosely-coupled algorithms that run as augmentation atop inertial navigation system (INS), which usually has a high rate processing of $100$ to $400$ Hz. 
In this paper, we adopt the loosely coupled \emph{Discorrelated minimum variance} (DMV) CL approach of~\cite{JZ-SSK-TRO:19} as our CL~framework and propose a neural network measurement scheduling approach for this algorithm. The agent's local measurement scheduler decides out of $n$ team members in its measurement zone, which ones the agent should take measurements from to gain the best localization improvement if it can only process $n_z< n$ number of measurements. Our proposed measurement scheduling is an approach that can be easily extended to any loosely-coupled CL scheme other than the DMV method. 

Measurement scheduling for computation/communication cost reduction has been studied in~\cite{chang2018optimal,mourikis2006optimal,caglioti2006cooperative,zhang2018multi,singh2017supermodulara,QY-LJ-SSK:20}\footnote{The interested reader can find sensor scheduling for other localization algorithms and state estimation also in~\cite{cieslewski2018data,tian2019resource,STJ-SLS:15}
among many others.}.
The method used can be divided into two categories: those that regulate the optimal measurement frequency given a fixed sensing graph throughout the whole operation~\cite{chang2018optimal,mourikis2006optimal}, 
and those that plan, at each timestep, what teammates a mobile agent should take measurement from when they are in its measurement zone~\cite{caglioti2006cooperative,zhang2018multi,singh2017supermodulara,QY-LJ-SSK:20}. Our problem of interest in this paper falls in the second category where the sensing graph of the team can change with time. 

Given a limited choices for measurement processing, the single-time-step optimal sensor scheduling aims to select those measurements that minimize the overall positioning uncertainty. However, the optimization problems for optimal measurement scheduling are known to be NP-hard~\cite{singh2017supermodulara,zhang2017sensor}. By establishing that the logdet of the joint covariance matrix of a tightly coupled joint CL is  submodular,~\cite{tzoumas2017scheduling,tzoumas2016nearoptimal,zhang2017sensor} propose sequential greedy algorithms (SGAs) for scheduling with known optimality gap. Another suboptimal greedy approach when the objective function is the trace of the joint covaraince matrix of the team is proposed in~\cite{singh2017supermodulara}. 
Even though~\cite{tzoumas2017scheduling,tzoumas2016nearoptimal,zhang2017sensor,singh2017supermodulara} offer suboptimal solutions with polynomial time complexity, they still suffer from high inter-agent communication, computation and memory cost to carry out the scheduling process. This is because, each agent is unaware of the other agents' local belief (local estimate and error covariance) without communication, making it impossible to perform measurement scheduling entirely locally. In a different approach,~\cite{QY-LJ-SSK:20} offers a greedy landmark selection heuristic that works based on minimizing an upper-bound on the total uncertainty of the team. However, this algorithm is restricted to a certain class of mobile agents moving on flat 2D terrain with the same motion model used in~\cite{chang2018optimal,mourikis2006optimal}.

\begin{figure}
    \centering
    \includegraphics[scale=0.15]{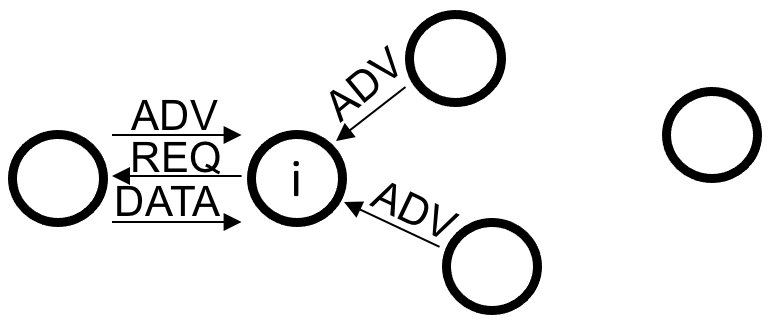}
    \caption{{\small In a data-driven SPIN protocol first a metadata (ADV)  is broadcast to announce the characteristic of the real data (DATA). Then,  DATA is only sent upon request (REQ).}}
    \label{fig::SPIN}
\end{figure}

\begin{figure}[t]
    \centering
    \includegraphics[trim=35pt 0pt 65pt 10pt,clip,scale = 0.3]{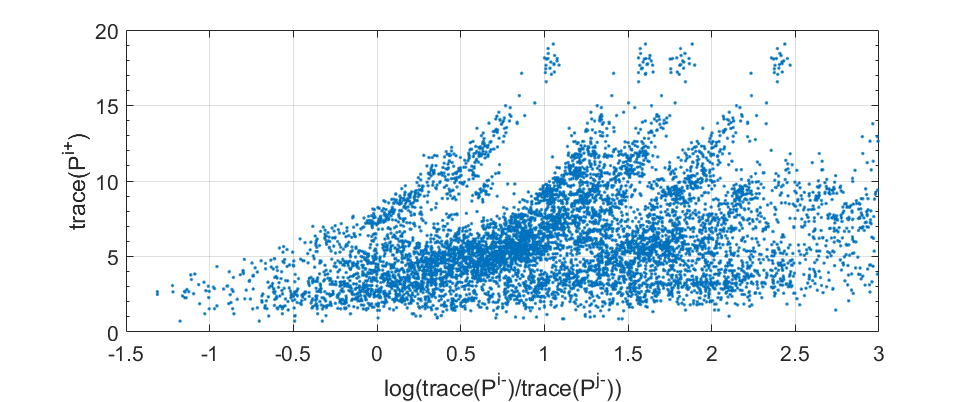}
    \caption{{\small The scatter plot of trace($\vect{P}^{i\updt}$) with respect to log(trace($\vect{P}^{i\prpg}$)/trace($\vect{P}^{j\prpg}$)) for numerical example of Section~\ref{sec::num}.}}
    \label{fig:tracescatter}
\end{figure}

The measurement scheduling in this paper builds on the results that show that the measurement scheduling for CL can be solved by a SGA algorithm. The novelty in our work is to use a neural network-based surrogate model, driven by local information and only a \emph{scalar} metadata from the landmark agents, as a proxy for the SGA's merit function. This solution addresses the time and memory complexity issues of running the SGA in three ways: (a) reducing the inter-agent communication message size, (b) decreasing
the complexity of function evaluations by using a simpler
surrogate (proxy) function, (c) reducing the required memory size. Inspired by the Sensor Protocol for Information via Negotiation (SPIN),  a data-centric dissemination protocol (see Fig.~\ref{fig::SPIN}), which we used in our earlier work for a negotiation-based communication protocol  for CL
~\cite{JZ-SSK:20sensor}, we use the trace of the covaraince of the candidate landmark agents as  a metadata to decide what landmark agents an agent should take measurement from. Unlike our work in~\cite{JZ-SSK:20sensor} which simply prioritized the landmark agents based on the size of their metadata, in this paper, we propose a measurement scheduling scheme that uses the metadata as an external input along with the local information of the agent to predict the CL update result. As shown in Fig.~\ref{fig:tracescatter}, the relationship between the ratio of local uncertainties and the updated covaraince is not a linear relation. To perform the prediction, we use a deep neural network as a data-driven solution. A multilayer perceptron (MLP) regression model is trained which outputs the trace of the updated covariance matrix.
Then, by implementing a sequential measurement update, the agent selects what other agents to take relative measurements from based on this prediction when they are only allowed to take limited measurements. The trained MLP has low complexity and is suitable to perform online prediction. The numerical results demonstrates the effectiveness of the proposed method. 

\section{Problem Definition}
\label{sec::prob}
\vspace{-0.08in}
Consider a group of $N$ mobile agents. The equation of motion of each agent $i\in\mathcal{V}=\{1,\cdots,N\}$ at timestep $t\in\mathbb{Z}^{+}$ is described~by
\begin{align}\label{eq::CL_sys_motion}
\vect{x}^i(t) &= \vect{f}(\vect{x}^i(t - 1),\vect{u}_m^i(t)),\quad \vect{x}^i\in\real^{n^i},	
\end{align}
where $\vect{x}^i(t)$ is the state of agent $i$ (position, attitude, velocity, etc) and
$\vect{u}^i_m = \vect{u}^i + \vect{\eta}^i \in\real^{m^i}$ is the dead-reckoning (IMU, wheel-encoders, or visual odometry) readout, with $\vect{u}^i$ as the actual value and $\vect{\eta}^i(t)\in\mathcal{N}(\vect{0}, \vect{Q}^i(t))$ the contaminating noise. The dead-reckoning readouts $\vect{u}_m^i(t)$ at each timestep  are used as input to propagate the posterior ego belief $\text{bel}^{i\updt}(t-1)=(\Hvect{x}^{i\updt}(t-1), \vect{P}^{i\updt}(t-1))$ which includes state estimate and the corresponding error covariance matrix $\vect{P}^{i\updt}\in\mathbb{S}^{++}_{n^i}$ according to the motion dynamics~\eqref{eq::CL_sys_motion} as
\begin{subequations}
\begin{align}
\Hvect{x}^{i\prpg}(t)&\!=\!f(\Hvect{x}^{i\updt}(t-1),\vect{u}_m^i(t)),\\
\vect{P}^{i\prpg}(t)&\!=\!\vect{F}^i(t)\vect{P}^{i\updt}(t-1)\vect{F}^i(t)^\top\!\!+\vect{G}^i(t)\vect{Q}^{i}(t)\vect{G}^i(t)^\top\!\!,
\end{align}
\end{subequations}
where $\vect{F}^{i}(t)=\nabla_{\vect{x}^i}\vect{f}|_{\textbf{x}^i=\Hvect{x}^{i\updt}(t-1),\vect{u}_m^i=\vect{0}}$ and $\textbf{G}^i(t)=\nabla_{\vect{\eta}^i}\vect{f}|_{\textbf{x}^i=\Hvect{x}^{i\updt}(t-1),\vect{u}^i=\vect{0}}$ are the Jacobian matrices. When there is no inter-agent measurement to update the local belief, we set $\text{bel}^{i\updt}(t)=\text{bel}^{i\prpg}(t)=(\Hvect{x}^{i\prpg}(t),\vect{P}^{i\prpg}(t))$, otherwise we~proceed to correct the local belief using DMV CL update approach~\cite{JZ-SSK-TRO:19} as outlined below.

Let the relative measurement (e.g., relative range, relative bearing, relative pose, or a combination of them) obtained by agent $i$ with respect to agent $j$ via agent $i$'s exteroceptive sensor be denoted by $i\rightarrow j$ and be modeled as
\begin{align}\label{eq::z_measurement}
    \vect{z}_j^i(t)=\vect{h}(\vect{x}^i(t),\vect{x}^j(t))+\vect{\nu}^i(t),\quad \vect{z}_j^i\in\real^{n_z^i},
\end{align}
where $\vect{\nu}^i(t)\in\mathcal{N}(\vect{0}, \vect{R}^i(t))$ is the zero mean Gaussian measurement noise with covaraince matrix $\vect{R}^i(t)$. Let $\Hvect{z}^i_{j}=\vect{h}(\Hvect{x}^{i\prpg},\Hvect{x}^{j\prpg})$, and $\vect{H}^i_{i}\!=\!\partial \vect{h}(\Hvect{x}^{i\prpg}\!,\Hvect{x}^{j\prpg})/\partial{\vect{x}}^i$ and $\vect{H}^i_{j}\!=\!\partial \vect{h}(\Hvect{x}^{i\prpg}\!,\Hvect{x}^{j\prpg})/\partial{\vect{x}}^j$ br the elements of the linearized model of $\vect{h}(\Hvect{x}^{i\prpg},\Hvect{x}^{j\prpg})$. DMV CL updates the propagated belief $\text{bel}^{i\prpg}(t)$ according~to
\footnote{DMV CL uses the discorrelated upper bound described below to account for the unknown cross-covaraince $\vect{P}_{ij}^{\prpg}(t)$ term of the joint covaraince matrix of  agents $i$ and $j$,
\!\!\!\!\begin{align*}
\!\!\begin{bmatrix*}
 \vect{P}^{i\prpg}(t)&\!\!\vect{P}_{ij}^{\prpg}(t)\\
 {\vect{P}_{ij}^{\prpg}}(t)^\top&\!\!\vect{P}^{j\prpg}(t)
 \end{bmatrix*}\!\leq\! \begin{bmatrix*}\frac{1}{\omega}\vect{P}^{i\prpg}(t)&\vect{0}\\\vect{0}&\!\!\frac{1}{1-\omega}\vect{P}^{j\prpg}(t)
\end{bmatrix*}
\!,~\omega\!\in\![0,1].
\end{align*}}
\begin{subequations}\label{eq::DMV}
\begin{align}
    \Hvect{x}^{i\updt}(t)&=\Hvect{x}^{i\prpg}(t)+\Bvect{\mathsf{K}}^i(\omega_\star^i)\,(\vect{z}^i_{j}(t)-\Hvect{z}^i_{j}(t)),\\
    \vect{P}^{i\updt}(t)&=\Bvect{\mathsf{P}}^{i}(\omega_\star^i),
\end{align}
\end{subequations}
where $ \Bvect{\mathsf{K}}^i(\omega)\!=\! \frac{\vect{P}^{i\prpg}}{\omega}{\vect{H}_i^i}^{\top}\big(\vect{H}_{i}^i\frac{\vect{P}^{i\prpg}}{\omega}{\vect{H}_{i}^i}^\top\!\!+\!\vect{H}_{j}^i\frac{\vect{P}^{j\prpg}}{1\!-\!\omega}{\vect{H}_{j}^i}\!^\top\!\!\!+\!\vect{R}^i\big)^{-1}\!\!\!,$
and 
\begin{align}
    \Bvect{\mathsf{P}}^{i}(\omega)=&\,
  \big(\,\omega(\vect{P}^{i\prpg})^{-1}+(1-\omega){{\vect{H}_{i}}^{i}}^{\top}({\vect{H}_{j}}^{i}\vect{P}^{j\prpg}{{\vect{H}_j}^{i}}^{\top}\nonumber\\&+(1-\omega)\vect{R}^{i})^{-1}{\vect{H}_i}^{i}\,\big)^{-1}\!.
  \label{eq::update_cov}
\end{align} 
The optimal $\omega$, denoted by $\omega^i_\star$, is obtained from  
\begin{align}\label{eq::omega_star}
\omega^i_\star=\underset{0\leq\omega\leq 1}{\text{argmin}} ~\log\det\,&\Bvect{\mathsf{P}}^{i}(\omega).
\end{align}

 To process multiple concurrent relative measurements $\{\vect{z}^i_l\}_{i\in\VV_z^i(t)}
 $ taken by agent $i$ from the set of agents $\VV_z^i(t)\subset\VV$ in its measurement zone at timestep $t$, we use sequential updating~\cite[page 103]{YBS-PKW-XT:11}. That is   $\text{bel}^{i\updt}(t)=(\Hvect{x}^{i\updt}(t,N_z),\Hvect{P}^{i\updt}(t,N_z))$ where for $a=\{1,\cdots N_z\}$, $N_z=|\VV_z^i(t)|$, we apply
\begin{align}\label{eq::sequential_update}(\Hvect{x}^{i\updt}(t,a),\Hvect{P}^{i\updt}(t,a))&\xleftarrow[]{Eq.~\eqref{eq::DMV}} \big(\Hvect{x}^{i\updt}(t,a-1),\\
&~\Hvect{P}^{i\updt}(t,a-1),\text{bel}^{v(a)\prpg}(t),{\vect{z}_{v(a)}}^{i}\big), \nonumber\end{align}
with $(\Hvect{x}^{i\updt}(t,0),\Hvect{P}^{i\updt}(t,0))=(\Hvect{x}^{i\prpg}(t),\Hvect{P}^{i\prpg}(t),\vect{z}_{v(1)}^i)$ and $v(a)$ being the $a$th element of $\VV_z^i(t)$. 

Per~\cite[Theorem 3.1]{JZ-SSK-TRO:19}, DMV update~\eqref{eq::DMV} using single measurement $\vect{z}_j^i$ satisfies $\det \vect{P}^{i\updt}(t)\leq \det \vect{P}^{i\prpg}(t)$. Thus, 
following the sequential updating scheme~\eqref{eq::sequential_update}, the more relative measurements are processed, the more improvement in estimation accuracy of agent $i$ due to anticipated reduction in total covariance can be achieved. However, to perform the update~\eqref{eq::sequential_update}, agent $i$ should acquire the local belief $\text{bel}^{l\prpg}(t)$ of every agent $l\in \VV_z^i(t)$, which comes at a cost of resources including energy and communication bandwidth. In addition, 
the time required to communicate and process the data reduces the propagation and updating rate that can lead to worsen localization performance. Measurement scheduling aims to make the best decision about what measurements to process when the resources limit the number of the relative measurements that can be processed. 

\textbf{Measurement Scheduling Problem (\sf{MeaSchProb})}:
Due to the constrained resources, only $q^i(t)\in\mathbb{Z}_{\geq 1}$ measurements can be processed at timestep $t$. Select a set of $q^i(t)$ agents, denoted by $\mathcal{V}^{i\updt}(t)$, from ${\mathcal{V}_z}^i(t)$ such that the updated estimate via sequential updating~\eqref{eq::sequential_update} using measurements from $\mathcal{V}^{i\updt}(t)$ results in maximum value for 
\begin{align}\label{eq::opt}
\!\!\!\underset{\mathcal{V}^{i\updt}(t)\subset \mathcal{V}_z^i(t)}{\text{max}}\!\!\!\!\! \text{trace}(\vect{P}^{i\prpg}(t)-\vect{P}^{i\updt}(t))~~\text{s.t.}~~|\mathcal{V}^{i\updt}(t)|\leq q^i(t).
\end{align}

{\sf{MeaSchProb}} is an NP-hard combinatorial optimization problem. Next, we propose a suboptimal solution to solve~\eqref{eq::opt} in polynomial time and with low communication~cost. 

\section{A neural network based distributed measurement scheduling}
\label{sec::mea_sche}
\vspace{-0.08in}
The updated covariance~\eqref{eq::update_cov} of agent $i$ using relative measurement $i\rightarrow j$, depends on its own  $\text{bel}^{i\prpg}$ and measurement
covariance $\vect{R}^{i}$ as well as $\text{bel}^{j\prpg}$ which is not available locally.  To perform CL update, the local belief of any candidate landmark agent $j\in\mathcal{V}_z^i(t)$ should be transmitted to agent $i$. However, communication of the $\text{bel}^{j\prpg}$ comes with high cost. In what follows, we propose a suboptimal solution to \textbf{\sf{MeaSchProb}} in which agent $i$ acquires only $\text{trace}(\vect{P}^{j\prpg}(t))$ of its candidate landmark agents $\mathcal{V}_z^i(t)$ as a metadata to use in a deep learning (DL) model
 to predict $\text{trace}(\vect{P}^{i\updt}(t))$ due to $i\to j$, $j\in \VV^i_z(t)$ without computing~\eqref{eq::update_cov}. 
 The communicated metadata $\text{trace}(\vect{P}^{i\updt}(t))$ in this process is a scalar, which has a much smaller communication message size than the entire belief of agent $j\in \VV^i_z(t)$.

\begin{figure}
    \centering
    \includegraphics[scale=0.29]{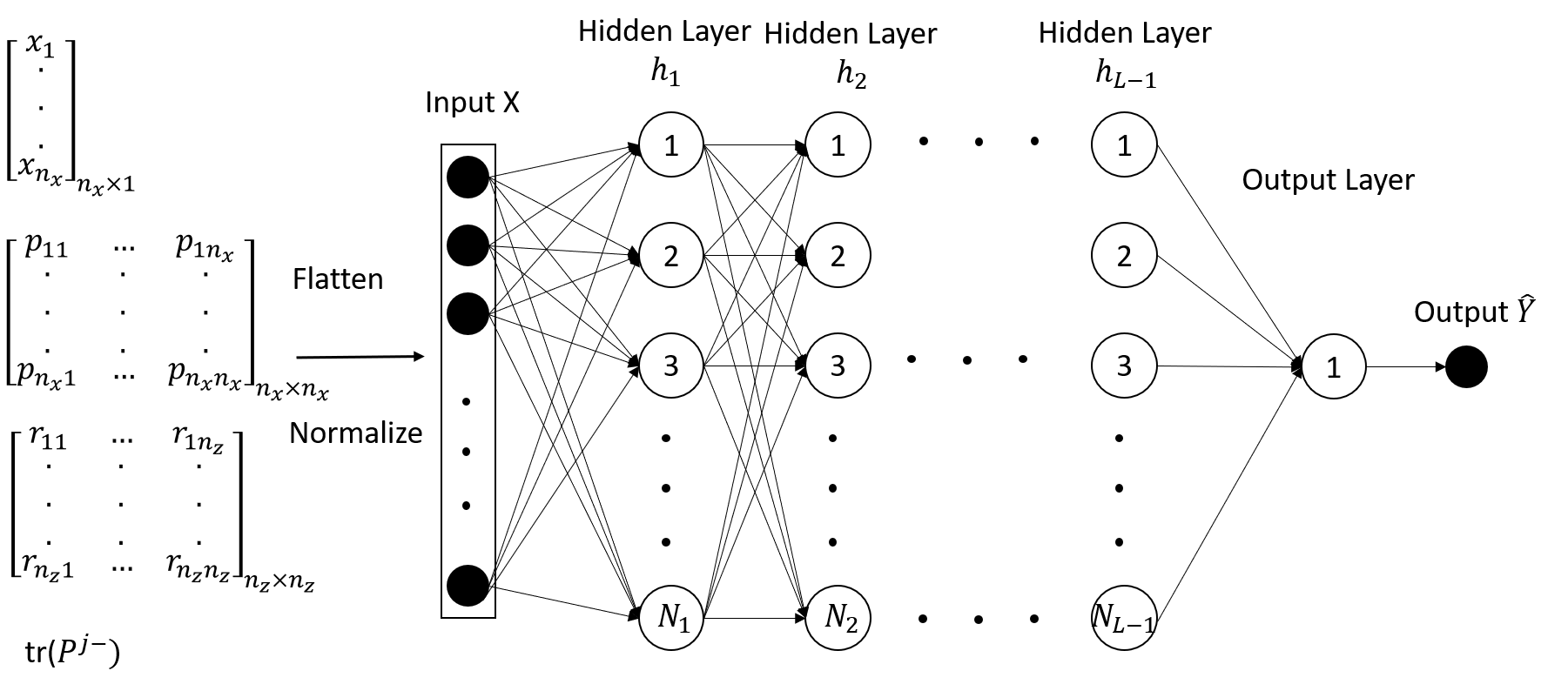}
    \caption{{\small The trace uncertainty prediction neural~network}}
    \label{fig:frame_work_dnn}
\end{figure}

With the available information 
$\text{bel}^{i\prpg}=(\Hvect{x}^{i\prpg}, \vect{P}^{i\prpg})$, measurement covariance $\vect{R}^{i}$, and  $\text{trace}(\vect{P}^{j\prpg})$, acquired through transmission, we flatten and normalize the data into an input vector, see Fig.~\ref{fig:frame_work_dnn}.  That is, the input and the corresponding label of each input is generated as
\begin{subequations}\label{eq::data_gen}
\begin{align}
    \vect{X} &= \text{\sf{generateInput}}(\Hvect{x}^{i\prpg}, \vect{P}^{i\prpg}, \vect{R}^{i}, \text{trace}(\vect{P}^{j\prpg})),\\
    Y &= \text{trace}(\vect{P}^{i\updt}).
\end{align}
\end{subequations}
Because the samples are labeled, we develop a supervised regression model using DNN to predict the updated $\hat{Y}$ without conducting the DMV CL update, see Fig.~\ref{fig:frame_work_dnn}, we denote this operation by
\begin{align}\label{eq::DNN}
\hat{Y}\xleftarrow[]{\text{DNN}}(\Hvect{x}^{i\prpg}, \vect{P}^{i\prpg}, \vect{R}^{i}, \text{trace}(\vect{P}^{j\prpg})).
\end{align}
DNN is used instead of the traditional ML model because of DNN's high performance given a large amount of data. Each pair of self-motion measurements and relative measurements could generate a set of training data in our setting. Due to the high sampling rate of sensors, a large amount of data could be collected and used for training. Even though DNN comes with high computational complexity, the forward propagation along the trained model is acceptable in real-time applications since the training process is finished off-line. The training and hyperparameter fine-tuning process for a DNN with the input and output data model as in~\eqref{eq::data_gen} for a numerical demonstration example is presented in Section~\ref{sec::num}. 

\subsection{A DNN-based distributed measurement scheduling}
\vspace{-0.05in}
DMV CL with the trained trace uncertainty prediction DNN~\eqref{eq::DNN} to perform the measurement scheduling follows a SG selection and sequential updating procedure as described in Algorithm~\ref{Alg::1}. Agent $i$ first obtains $\text{trace}(\vect{P}^{j\prpg}(t))$, $j\in \VV^i_z(t)$. Next, the scheduler routine starts by $\VV^i_z(t)$ and iteratively chooses the landmark agent $j$ that has the maximum $\hat{Y}$ value, predicted by the DNN, among all in $\VV^i_z(t)$, obtains agent $j$'s local belief via communication, and updates the agent $i$'s local belief using the DMV procedure~\eqref{eq::DMV} using relative measurement from $j$. The scheduler routine updates $\VV^i_z(t)\leftarrow \VV^i_z(t)\backslash\{j\}$ and continues the sequential selection/update procedure until $q^i(t)$ landmark agents are chosen.  Let $|\VV_z^i(t)| = m$, and for simplicity of presentation let $n^j=n$ and $n_z^j={n_z}$ (recall~\eqref{eq::CL_sys_motion} and~\eqref{eq::z_measurement}) for all agents $j\in\VV_z^i(t) $. Table~\ref{table:complexity} summarizes how computation, memory and communication complexity of Algorithm~\ref{Alg::1} compares to those of SGA's. In Algorithm~\ref{Alg::1}, instead of requesting the local belief of every landmark agent $j\in\VV_z^i(t)$ first a scalar meta data of landmark agents $j\in\VV_z^i(t)$ is requested and then only the belief of $q^i$ agents selected by the scheduler is requested one by one (reducing the memory usage by overwriting). The computational difference of the DNN-based and SG measurement scheduling is in line 6 of Algorithm~\ref{Alg::1}. The computational complexity of $\hat{Y}_j$ is $O(\sum^{L-1}_{l=0}\mathsf{n
}_{(l)}\mathsf{n}_{(l+1)})$ where $\mathsf{n}_{(l)}$ is the number of units in the $l^{\text{th}}$ hidden layer and $L$ is the number of hidden layers. In SGA instead of $\hat{Y}_j$, first $\vect{P}^{\updt i}$ using equations~\eqref{eq::update_cov} and solving the optimization problem~\eqref{eq::omega_star} for $\omega_\star^i$ should be computed and then $\vect{Y}=\text{trace}(\vect{P}^{i\updt})$. The computational complexity of computing  $\log\det\,\Bvect{\mathsf{P}}^{i}(\omega)$ with matrix inversion and multiplication operations is $O({n^3}+{n_z^2}{n^2})$. Assuming that a line search algorithm with $M$ steps are used to solve~\eqref{eq::omega_star}, the computational complexity of $\vect{Y}$ is of order $O(M\,(n^3+{{n_z}^2}{n^2}))$.

\begin{algorithm}[t]
\caption{{\small DNN-based measurement scheduling and update procedure}  }
\label{Alg::1}
{\footnotesize
\begin{algorithmic}[1]
\State $\mathbf{\textbf{Inputs:}}~ \Hvect{x}^{i\prpg}, \vect{P}^{i\prpg}, \vect{R}^{i}, \VV^i_z(t)$ 
\State Obtain $\text{trace}(\vect{P}^{j\prpg})$ from every agent $j\in \VV^i_z(t)$
\State $l\leftarrow 1$
\While {$l\leq q^i(t)$}
 \For{$j\in\mathcal{V}_z^i(t)$}
        \State $\hat{Y}_j\xleftarrow[]{\text{DNN}}(\Hvect{x}^{i\prpg}, \vect{P}^{i\prpg}, \vect{R}^{i}, \text{trace}(\vect{P}^{j\prpg}))$
      \EndFor
      \State $j^\star=\argmax_{j\in\mathcal{V}_z^i(t) } \hat{Y}_j$
      \State Obtain $\text{bel}^{j^\star\prpg}(t)$ and $\vect{z}_{j^\star}^i$ from agent $j^\star$
      \State $(\Hvect{x}^{i\prpg}, \vect{P}^{i\prpg})\xleftarrow[]{Eq.~\eqref{eq::DMV}} \big(\Hvect{x}^{i\prpg}, \vect{P}^{i\prpg},\text{bel}^{j^\star\prpg}(t),\vect{z}_{j^\star}^i\big)$
      \State $\mathcal{V}_z^i(t)\leftarrow\mathcal{V}_z^i(t)\backslash\{j^\star\}$
      \State $l\leftarrow l+1$
\EndWhile
\State $(\Hvect{x}^{i\updt}, \vect{P}^{i\updt})\leftarrow \big(\Hvect{x}^{i\prpg}, \vect{P}^{i\prpg}\big)$
\State \textbf{Return} $(\Hvect{x}^{i\updt}, \vect{P}^{i\updt})$
\end{algorithmic}
}
\end{algorithm} 
\begin{table}[t]
\centering
\caption{{\small Complexity analysis for SG algorithms and proposed learning based algorithm for solving MeaScheProb in~\eqref{eq::opt}}}.\label{table:complexity}
{\scriptsize
\begin{tabular}{cccc}
\hlinewd{1.5pt}
  \!\!\!   & \!\!\!\!\!\!Computation & \!\!\!\!\!\!Memory&\!\!\!\!\!\!communication\\
\hline
\!\!\!  \!\!\!\!  SGA &\!\!\!\!\!\!\!\! $O(m(m\!-\!q^i)M(n^3+n_z^2n^2))$ &\!\!\!\!\!\!\!\!$O(m n^2)$&\!\!\!\!$O(m n^2)$\\
\!\!\!  \!\!\!\!  Algorithm~\ref{Alg::1} &\!\!\!\!\!\!\!\! $O(m(m-q^i)\sum^{L-1}_{l=0}\mathsf{n}_{(l)}\mathsf{n}_{(l+1)})$ &\!\! \!\!\!\!$O(n^2\! +\! m)$&\!\!\!\!\!\!$O(q^i \,n^2 \!+\! m)$\\
\hlinewd{1.5pt}
\end{tabular}
}
\vspace{-0.2in}
\end{table}

\section{Numerical Result}\label{sec::num}
\vspace{-0.08in}
We demonstrate the performance of our proposed scheduling method in a simulation study using the public experimental UTIAS multi-robot CL and mapping dataset for $5$ robots moving on a 2D flat surface in an indoor environment~\cite{leung2011utias}. The UTIAS dataset consists of 9 sub-datasets, and each includes measurement data, odometry data, and ground truth position data of all the team members. The motion of each robot is described by,  $x^i(t+1)\!=\!x^i(t)+\Delta t (v_m^i(t)\cos(\phi^i(t)))$,
    $y^i(t+1)\!=\!x^i(t)+\Delta t (v_m^i(t)\sin(\phi^i(t)))$,
    $\phi^i(t+1)\!=\!\phi^i(t)+\Delta t \,\omega_m^i(t)$, $i\in\{1,2,3\}$,
where $\vect{x}^i=[x^i, y^i, \phi^i]^\top$.  
Here $v_m^i(t)=v^i(t)+\nu_v^i(t)$ and $\omega^i_m(t)=\omega^i(t)+\nu_{\omega}^i(t)$ are measured linear and angular velocities, while $\nu^i_v$ and $\nu^i_\omega$ are the corresponding contaminating measurement noises with variance $\sigma_{\nu^i}=2.253|v_m^i(t)|~m/s$ and $\sigma_{\omega^i}=0.587~rad/s$. The actual velocities are $v^i(t)$ and $\omega^i(t)$. The inter-agent measurements are relative range and relative bearing described by
$
\vect{h}(\vect{x}^i,\vect{x}^j)=\left[\begin{smallmatrix}
\sqrt{(x^i-x^j)^2 + (y^i-y^j)^2}\\
\phi^j - \phi^i
\end{smallmatrix}\right]+\left[\begin{smallmatrix}
\nu^i_{\rho}\\
\nu^i_{\theta}
\end{smallmatrix}\right]
$, where $\nu^i_{\rho}$ and $\nu^i_{\theta}$ are measurement noises whose variances are, respectively, $\sigma_{\rho^i}=0.147~m$ and $\sigma_{\theta^i}=0.1 ~rad$.

\subsection{Trace uncertainty prediction neural network}\label{sec::training}
\vspace{-0.05in}
To generate independent and identically distributed samples for training, developing, and testing, we perform single-step propagation and update from initial belief with randomly generated errors for each pair of self-motion measurement and relative measurement. From the public dataset, we generated $50987$ samples which is split into training set ($40987$ samples), developing set ($5000$ samples), and test set ($5000$ samples). The data is preprocessed. The vector and matrices in input $\vect{X}$ is flattened to a vector with 16 features. The features in the input are normalized to achieve easier optimization and faster~learning. 

We use an L-layer neural network as shown in Fig.~\ref{fig:frame_work_dnn}. We have $16$ features for this particular dataset in the input layer and only $1$ unit in the output layer since the output of the model is only a scalar. The layers are fully connected. The number of hidden layers and the number of units in each layer is fine-tuned during training. The activation function is Rectified Linear Unit (ReLU)~\cite{hara2015analysis} for each hidden layer and linear function for output layer. To evaluate the performance, we use Mean-squared-error (MSE) as our loss function due to its convexity and thus the ease to optimize using backpropagation method
\begin{align}
    \mathcal{L}(Y,\hat{Y})=||Y - \hat{Y}||^{2}.
    \label{eq::loss}
\end{align}
Therefore, the training process is defined as
\begin{equation}
    \underset{\vect{W}}{\min}J(\vect{W})=\underset{\vect{W}}{\min}\frac{1}{m}\sum\nolimits_{k=1}^{m}\mathcal{L}(Y^{(k)},\hat{Y}^{(k)}),
\end{equation}
where $m$ is the total number of training samples. 

 To train and test the model, we focus on tuning the hyperparameters, including the number of hidden layers $L$, the number of hidden units in each layer $\mathsf{n}_{(l)}$, the learning rate $\alpha$, and the number of epochs. Since the hyperparameters cannot be determined analytically, we fine-tune the hyperparameters through repeated experimentations, which makes the role of the  training speed extremely important. To accelerate the training process, adaptive moment estimation (Adam)~\cite{kingma2014adam} is used. Adam combines RMSprop and Gradient Descent~(GD) with momentum. It uses the squared gradients to scale the learning rate as RMSprop, and it takes advantage of momentum by using the moving average of the gradient instead of gradient as GD with momentum. Due to the large training set size, we use the mini-batch method, which calculates the error and updates the parameters for each batch instead of the whole training set using a batch size of $256$. 

\begin{figure}
    \centering
    \includegraphics[trim=70pt 20pt 10pt 30pt, clip,scale=0.278]{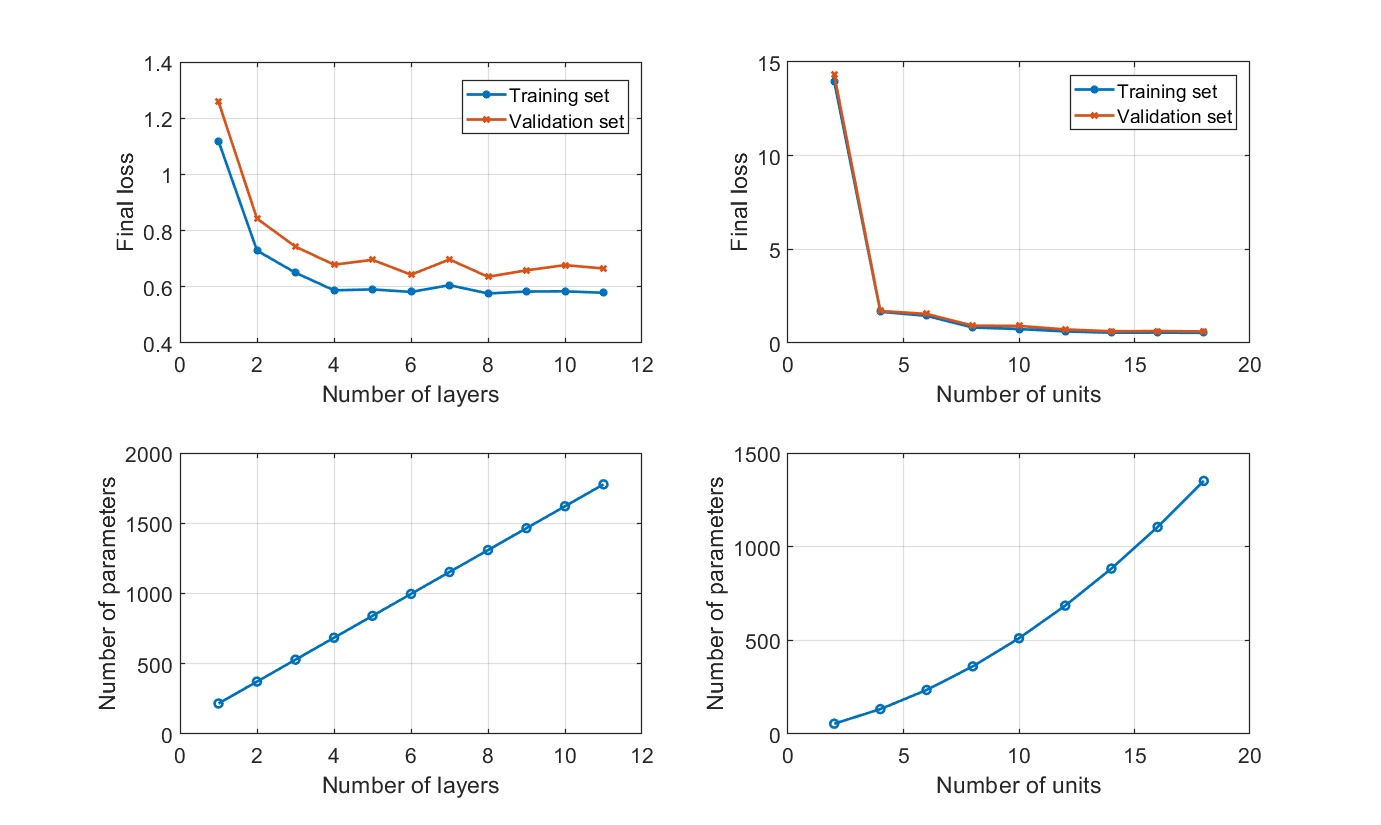}
    \caption{{\small The final training loss of DNN model and the corresponding number of parameters (bottom) with varying number of hidden layers (left) and varying number of hidden units in each layer (right).}}
    \label{fig:training_size}\vspace{-0.05in}
\end{figure}

\begin{figure}[t]
    \centering
        \includegraphics[trim=10pt 5pt 0 30pt,scale=0.29]{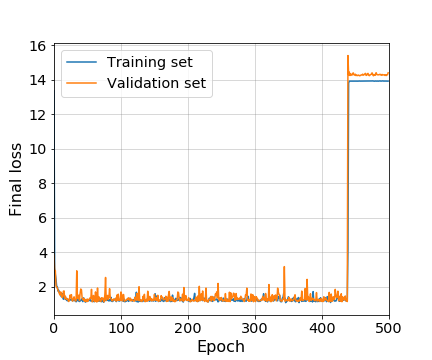}
 \!\!       \includegraphics[trim=10pt 5pt 10pt 30pt,scale=0.29]{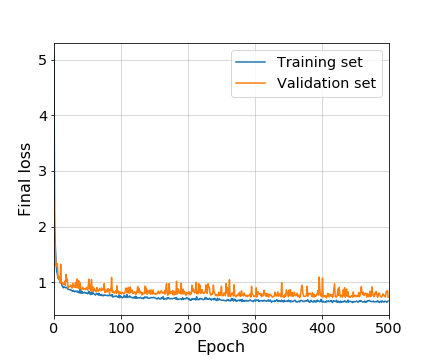}
    \caption{{\small Training with learning rate of $0.1$ (left) and $0.01$ (right).}}\vspace{-0.13in}
    \label{fig:training_rate}
\end{figure}

 During the tuning process,  we vary and tune the hyperparameters one at a time. 
 Even though the tuning result is not guaranteed to be optimal, good performance and high efficiency are achieved. We start with tuning the number of layers. We set 12 hidden units in each layer, $0.02$ learning rate, and run $500$ epochs for each model with a different number of layers from 1 hidden layer to 11 hidden layers. The result is shown in Fig.~\ref{fig:training_size}(left). The final training loss decreases dramatically from a really simple single-layer model and converges at 4 layers. As the neural network goes deeper, the number of parameters grows in Fig.~\ref{fig:training_size}(left-bottom) and leads to higher computational complexity for real-time implementation on measurement scheduling. So, we set the number of hidden layers to $4$. In the second, we set 4 hidden layers, $0.02$ learning rate, and run $500$ epochs for each model with a different number of hidden units from 2 hidden layers to 18 hidden layers. The result in Fig.~\ref{fig:training_size}(right) converges at 14 hidden units and keeps increasing the size does not improve the performance anymore.
 Learning rate $\alpha$ affects how much to change the model in response to the loss. Small training rate may result in requiring more epochs to converge and has a chance to stuck at the saddle point. A large learning rate may converge quickly to a local minimum instead of the global one. By varying the learning rate from $1$ to $0$ through multiple testing, we set the optimal learning rate to be 0.01. As in Fig.~\ref{fig:training_rate}(left), the training converges to a local minimum due to a high learning rate at 0.1. Fig.~\ref{fig:training_rate}(right) is the learning process with $\alpha = 0.01$. As we can see, the loss converges quickly and converges at around 200 epochs. After fine-tuning the DNN model, we finalize the design of the model as in Table.~\ref{tab:my_label}.

\begin{table}[t]
    \caption{{\small Parameters of the trace uncertainty prediction DNN.}}
    \label{tab:my_label}
    \centering
\begin{tabular}{l l}
\hlinewd{1.5pt}
     Hyperparameters & Value\\
     \hline
     Number of hidden layers $L$ & 4\\
     Number of units $\mathsf{n}_{(l)}$ per layer & 14\\ Loss & MSE\\ 
     Activation function & ReLU \\
     Optimizer & ADAM \\ Batch size & 256\\
     Learning rate $\alpha$ & 0.01 \\Epochs  & 200\\
\hlinewd{1.5pt}
\end{tabular}
\end{table}

\subsection{Numerical Evaluation}
\begin{figure}[t]
    \centering
    \includegraphics[trim=70pt 0 10pt 10pt,clip,scale = 0.26]{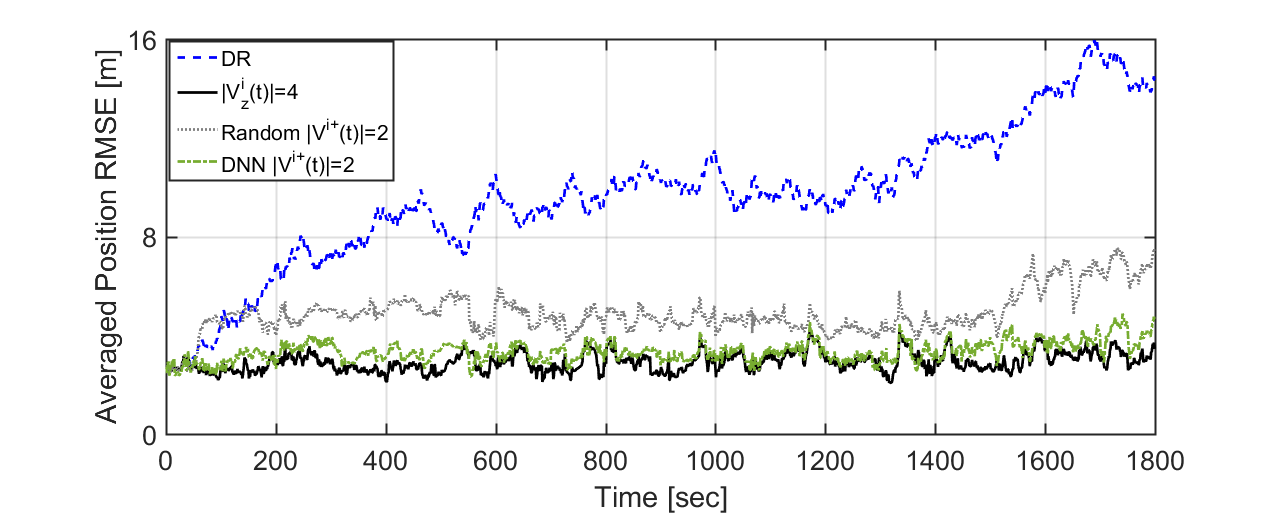}
    \caption{{\small The averaged position RMSE for the numerical study.}}\vspace{-0.05in}
    \label{fig:exp_result_numerical}
\end{figure}

\begin{figure}[t]
    \centering
    \includegraphics[trim=5pt 0pt 3pt 1pt,clip,width =0.44 \textwidth]{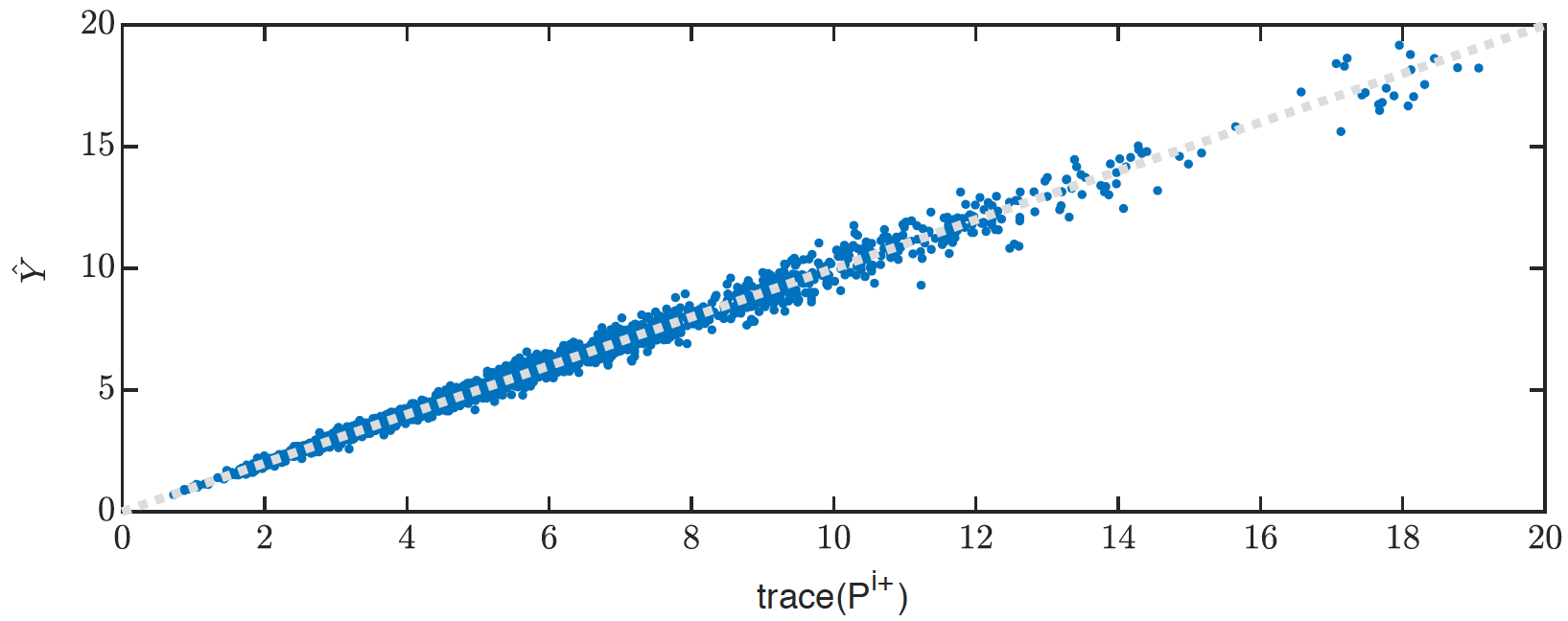}
    \caption{{\small The scatter plot of the predicted trace of updated covariance ($\hat{Y}$) vs. the actual trace of updated covariance. Note that the points are scattered closely around the line with slope $1$.}}
    \label{fig:prediction}
\end{figure}

\vspace{-0.05in}
We evaluate the efficiency of our proposed scheduling algorithm by comparing its localization performance in a resource constrained scenario with the performance of a random selection  (select measurements randomly). To avoid over-optimism, we use a different subsets of UTIAS data from the data used for model training. 
In our simulation scenarios, at each timestep, each agent can take at most 4 inter-agent measurements from its teammates in this setting.
The average position root-mean-squared-error (RMSE) over the 5 robots are shown in Fig.~\ref{fig:exp_result_numerical}.
Processing all the inter-agents measurements results in the best localization accuracy as the black solid plot shows in Fig~\ref{fig:exp_result_numerical}. However, under resource constraint, not all the measurements can be scheduled to be processed. Let the maximum number of measurements that can be processed be $q^i=2$. If we simply select 2 measurements randomly from 4, the result is shown as the gray dotted plot in Fig~\ref{fig:exp_result_numerical}, which is clearly inferior to measurement accuracy achieved when our proposed proposed DNN based measurement selection method is implemented (green plot in Fig~\ref{fig:exp_result_numerical}). Figure~\ref{fig:prediction} shows the predicted trace of updated covariance, $\hat{Y}$, vs. the actual trace of the updated covariance of this simulation study, which showcases the satisfactory performance of our designed DNN~predictor.

 \section{Conclusions}
 \label{sec::con}
 \vspace{-0.08in}
 In this paper, we addressed  the measurement scheduling problem in loosely-coupled cooperative localization. We developed a neural network regression based distributed measurement scheduling that only requires scalar-wise inter-agent information exchange. The model used only the locally available information of the agent that wanted to perform the scheduling and the trace of the covariance matrices of its collaborating team mates to predict the trace of the updated covariance of the scheduling agent. Then, the updates with the most prospective improvement are scheduled if there are only a limited number of inter-agent measurements allowed at that given time instant. This method fits the framework of our negotiation-based rescheduling communication protocols. A simulation study demonstrated our results.

\bibliographystyle{IEEEtran}
\bibliography{IEEEabrv,my_ref.bib,ref_schdul.bib}

\begin{thebibliography}{10}
\providecommand{\url}[1]{#1}
\csname url@samestyle\endcsname
\providecommand{\newblock}{\relax}
\providecommand{\bibinfo}[2]{#2}
\providecommand{\BIBentrySTDinterwordspacing}{\spaceskip=0pt\relax}
\providecommand{\BIBentryALTinterwordstretchfactor}{4}
\providecommand{\BIBentryALTinterwordspacing}{\spaceskip=\fontdimen2\font plus
\BIBentryALTinterwordstretchfactor\fontdimen3\font minus
  \fontdimen4\font\relax}
\providecommand{\BIBforeignlanguage}[2]{{%
\expandafter\ifx\csname l@#1\endcsname\relax
\typeout{** WARNING: IEEEtran.bst: No hyphenation pattern has been}%
\typeout{** loaded for the language `#1'. Using the pattern for}%
\typeout{** the default language instead.}%
\else
\language=\csname l@#1\endcsname
\fi
#2}}
\providecommand{\BIBdecl}{\relax}
\BIBdecl

\bibitem{SEW-JMW-LLW-RME:13}
S.~E. Webster, J.~M. Walls, L.~L. Whitcomb, and R.~M. Eustice, ``Decentralized
  extended information filter for single-beacon cooperative acoustic
  navigation: theory and experiments,'' \emph{IEEE Transactions on Robotics},
  vol.~29, no.~4, pp. 957--974, 2013.

\bibitem{bahr2009consistent}
A.~Bahr, M.~R. Walter, and J.~J. Leonard, ``Consistent cooperative
  localization,'' in \emph{2009 {{IEEE International Conference}} on
  {{Robotics}} and {{Automation}}}, 2009, pp. 3415--3422.

\bibitem{JN-JR-PH-IS-MO-KVSH:14}
J.~. {Nilsson}, J.~{Rantakokko}, P.~{Händel}, I.~{Skog}, M.~{Ohlsson}, and
  K.~V.~S. {Hari}, ``Accurate indoor positioning of firefighters using dual
  foot-mounted inertial sensors and inter-agent ranging,'' in \emph{2014
  IEEE/ION Position, Location and Navigation Symposium - PLANS 2014}, May 2014,
  pp. 631--636.

\bibitem{zhu2018loosely}
J.~Zhu and S.~S. Kia, ``A loosely coupled cooperative localization augmentation
  to improve human geolocation in indoor environments,'' in \emph{International
  Conference on Indoor Positioning and Indoor Navigation}.\hskip 1em plus 0.5em
  minus 0.4em\relax IEEE, 2018, pp. 206--212.

\bibitem{SSK-SF-SM:16}
S.~S. Kia, S.~Rounds, and S.~Mart{\'\i}nez, ``Cooperative localization for
  mobile agents: a recursive decentralized algorithm based on {K}alman filter
  decoupling,'' \emph{{IEEE} Control Systems Magazine}, vol.~36, no.~2, pp.
  86--101, 2016.

\bibitem{roumeliotis2002distributed}
S.~I. Roumeliotis and G.~A. Bekey, ``Distributed multirobot localization,''
  \emph{IEEE transactions on robotics and automation}, vol.~18, no.~5, pp.
  781--795, 2002.

\bibitem{carrillo-arce2013decentralized}
L.~C. Carrillo-Arce, E.~D. Nerurkar, J.~L. Gordillo, and S.~I. Roumeliotis,
  ``Decentralized multi-robot cooperative localization using covariance
  intersection,'' in \emph{2013 {{IEEE}}/{{RSJ International Conference}} on
  {{Intelligent Robots}} and {{Systems}}}, 2013, pp. 1412--1417.

\bibitem{kia2018serverassisted}
S.~S. Kia, J.~Hechtbauer, D.~Gogokhiya, and S.~Mart{\'\i}nez, ``Server-assisted
  distributed cooperative localization over unreliable communication links,''
  \emph{IEEE Transactions on Robotics}, vol.~34, no.~5, pp. 1392--1399, 2018.

\bibitem{POA-CR-RKM:01}
P.~O. Arambel, C.~Rago, and R.~K. Mehra, ``Covariance intersection algorithm
  for distributed spacecraft state estimation,'' in \emph{{A}merican {C}ontrol
  {C}onference}, Arlington, Virginia, USA, 2001, pp. 4398--4403.

\bibitem{HL-FN:13}
H.~Li and F.~Nashashibi, ``Cooperative multi-vehicle localization using split
  covariance intersection filter,'' \emph{IEEE Intelligent Transportation
  Systems Magazine}, vol.~5, no.~2, pp. 33--44, 2013.

\bibitem{DM-NO-VC:13}
D.~Marinescu, N.~O'Hara, and V.~Cahill, ``Data incest in cooperative
  localisation with the common past-invariant ensemble kalman filter,'' in
  \emph{IEEE Int. Conf. on Information Fusion}, {I}stanbul, {T}urkey, 2013, pp.
  68--76.

\bibitem{JZ-SSK-TRO:19}
J.~Zhu and S.~S. Kia, ``Cooperative localization under limited connectivity,''
  \emph{IEEE Transactions on Robotics}, vol.~35, no.~6, pp. 1523--1530, 2019.

\bibitem{LCC-EDN-JLG-SIR:13}
L.~C. Carrillo-Arce, E.~D. Nerurkar, J.~L. Gordillo, and S.~I. Roumeliotis,
  ``Decentralized multi-robot cooperative localization using covariance
  intersection,'' in \emph{IEEE/RSJ Int. Conf. on Intelligent Robots and
  Systems}, Tokyo, Japan, 2013, pp. 1412--1417.

\bibitem{JSR-MY-BDOA-HH-PS:19}
J.~S. Russell, M.~Ye, B.~D.~O. Anderson, H.~Hmam, and P.~Sarunic, ``Cooperative
  localization of a gps-denied uav using direction-of-arrival measurements,''
  \emph{{I}{E}{E}{E} {T}ransactions on {A}erospace and {E}lectronic {S}ystems},
  vol.~56, no.~3, pp. 1966 -- 1978, 2019.

\bibitem{chang2018optimal}
T.-K. Chang and A.~Mehta, ``Optimal scheduling for resource-constrained
  multirobot cooperative localization,'' \emph{IEEE Robotics and Automation
  Letters}, vol.~3, no.~3, pp. 1552--1559, 2018.

\bibitem{mourikis2006optimal}
A.~I. Mourikis and S.~I. Roumeliotis, ``Optimal sensor scheduling for
  resource-constrained localization of mobile robot formations,'' \emph{IEEE
  Transactions on Robotics}, vol.~22, no.~5, pp. 917--931, 2006.

\bibitem{caglioti2006cooperative}
V.~Caglioti, A.~Citterio, and A.~Fossati, ``Cooperative, distributed
  localization in multi-robot systems: a minimum-entropy approach,'' in
  \emph{IEEE Workshop on Distributed Intelligent Systems: Collective
  Intelligence and Its Applications}, Prague, Czech Republic, 2006, pp. 25--30.

\bibitem{zhang2018multi}
L.~Zhang, X.~Tao, and H.~Liang, ``Multi {AUV}s cooperative navigation based on
  information entropy,'' in \emph{MTS/IEEE OCEANS}, Charleston, South Carolina,
  USA, 2018, pp. 1--10.

\bibitem{singh2017supermodulara}
P.~Singh, M.~Chen, L.~Carlone, S.~Karaman, E.~Frazzoli, and D.~Hsu,
  ``Supermodular mean squared error minimization for sensor scheduling in
  optimal {{Kalman Filtering}},'' in \emph{American Control Conference},
  Seattle, Washington, USA, 2017, pp. 5787--5794.

\bibitem{QY-LJ-SSK:20}
Q.~Yan, L.~Jiang, and S.~S. Kia, ``Measurement scheduling for cooperative
  localization in resource-constrained conditions,'' \emph{{I}{E}{E}{E}
  {R}obotics and {A}utomation {L}etters}, vol.~5, no.~2, pp. 1991--1998, 2020.

\bibitem{cieslewski2018data}
T.~{Cieslewski}, S.~{Choudhary}, and D.~{Scaramuzza}, ``Data-efficient
  decentralized visual slam,'' in \emph{2018 IEEE International Conference on
  Robotics and Automation (ICRA)}, May 2018, pp. 2466--2473.

\bibitem{tian2019resource}
Y.~Tian, K.~Khosoussi, and J.~P. How, ``A resource-aware approach to
  collaborative loop closure detection with provable performance guarantees,''
  \emph{The International Journal of Robotics Research}, 2019.

\bibitem{STJ-SLS:15}
S.~T. Jawaid and S.~L. Smith, ``Submodularity and greedy algorithms in sensor
  scheduling for linear dynamical systems,'' \emph{{Automatica}}, vol.~61, pp.
  282--288, 2015.

\bibitem{zhang2017sensor}
H.~Zhang, R.~Ayoub, and S.~Sundaram, ``Sensor selection for kalman filtering of
  linear dynamical systems: Complexity, limitations and greedy algorithms,''
  \emph{Automatica}, vol.~78, pp. 202--210, 2017.

\bibitem{tzoumas2017scheduling}
V.~Tzoumas, N.~A. Atanasov, A.~Jadbabaie, and G.~J. Pappas, ``Scheduling
  nonlinear sensors for stochastic process estimation,'' in \emph{American
  Control Conference}.\hskip 1em plus 0.5em minus 0.4em\relax IEEE, 2017, pp.
  580--585.

\bibitem{tzoumas2016nearoptimal}
V.~Tzoumas, A.~Jadbabaie, and G.~J. Pappas, ``Near-optimal sensor scheduling
  for batch state estimation: Complexity, algorithms, and limits,'' in
  \emph{2016 IEEE 55th Conference on Decision and Control}.\hskip 1em plus
  0.5em minus 0.4em\relax IEEE, 2016, pp. 2695--2702.

\bibitem{JZ-SSK:20sensor}
J.~Zhu and S.~S. Kia, ``A spin-based dynamic {T}{D}{M}{A} communication for a
  {U}{W}{B}-based infrastructure-free cooperative navigation,'' \emph{IEEE
  Sensors Letters}, vol.~4, no.~7, pp. 1--4, 2020.

\bibitem{YBS-PKW-XT:11}
Y.~Bar-Shalom, P.~K. Willett, and X.~Tian, \emph{Tracking and Data Fusion: A
  Handbook of Algorithms}.\hskip 1em plus 0.5em minus 0.4em\relax Taylor \&
  Francis, 2011.

\bibitem{leung2011utias}
K.~Y. Leung, Y.~Halpern, T.~D. Barfoot, and H.~H. Liu, ``The {U}{T}{I}{A}{S}
  multi-robot cooperative localization and mapping dataset,'' \emph{The
  International Journal of Robotics Research}, vol.~30, no.~8, pp. 969--974,
  2011.

\bibitem{hara2015analysis}
K.~Hara, D.~Saito, and H.~Shouno, ``Analysis of function of rectified linear
  unit used in deep learning,'' in \emph{International Joint Conference on
  Neural Networks (IJCNN)}, Killarney, Ireland, 2015, pp. 1--8.

\bibitem{kingma2014adam}
D.~P. Kingma and J.~Ba, ``Adam: A method for stochastic optimization,''
  \emph{arXiv preprint arXiv:1412.6980}, 2014.

\end{thebibliography}

\end{document}